\documentclass[conference]{IEEEtran}
\IEEEoverridecommandlockouts

\usepackage{cite}
\usepackage{amsmath,amssymb,amsfonts}
\usepackage{algorithmic}
\usepackage{graphicx}
\usepackage{textcomp}
\usepackage{blindtext}
\usepackage{hyperref}
\usepackage{xcolor}
\usepackage{booktabs}
\usepackage{multirow}
\usepackage{threeparttable} 
\usepackage{tabularx}

\def\BibTeX{{\rm B\kern-.05em{\sc i\kern-.025em b}\kern-.08em
    T\kern-.1667em\lower.7ex\hbox{E}\kern-.125emX}}
\begin{document}

\title{MMREC: LLM Based Multi-Modal Recommender System \\}
\IEEEpeerreviewmaketitle

\author{\IEEEauthorblockN{1\textsuperscript{st} Jiahao Tian}
\IEEEauthorblockA{
\textit{Georgia Institute of Technology}\\
Atlanta, Georgia, USA \\
jtian83@gatech.edu}
\and
\IEEEauthorblockN{2\textsuperscript{nd} Zhenkai Wang}
\IEEEauthorblockA{
\textit{The University of Texas at Austin}\\
Austin, Texas, USA \\
kay.zhenkai.wang@utexas.edu}
\and
\IEEEauthorblockN{3\textsuperscript{rd} Jinman Zhao}
\IEEEauthorblockA{
\textit{University of Toronto}\\
Toronto, Ontario, Canada \\
jinman.zhao@mail.utoronto.ca}
\and
\IEEEauthorblockN{4\textsuperscript{th} Zhicheng Ding}
\IEEEauthorblockA{\textit{Columbia University} \\
New York, NY, USA \\
zhicheng.ding@columbia.edu}
}
\maketitle

\begin{abstract}
The importance of recommender systems is growing rapidly due to the exponential increase in the volume of content generated daily. This surge in content presents unique challenges for designing effective recommender systems. Key among these challenges is the need to effectively leverage the vast amounts of natural language data and images that represent user preferences. 
This paper presents a novel approach to enhancing recommender systems by leveraging Large Language Models (LLMs) and deep learning techniques. The proposed framework aims to improve the accuracy and relevance of recommendations by incorporating multi-modal information processing and by the use of unified latent space representation. The study explores the potential of LLMs to better understand and utilize natural language data in recommendation contexts, addressing the limitations of previous methods. The framework efficiently extracts and integrates text and image information through LLMs, unifying diverse modalities in a latent space to simplify the learning process for the ranking model. Experimental results demonstrate the enhanced discriminative power of the model when utilizing multi-modal information. This research contributes to the evolving field of recommender systems by showcasing the potential of LLMs and multi-modal data integration to create more personalized and contextually relevant recommendations.
\end{abstract}


\section{Introduction}
Recommender Systems (RS)  have become an integral component of modern digital ecosystems, playing a pivotal role in personalizing user experiences across various domains such as e-commerce, streaming services and more \cite{guo2017deepfm, zhou2018deep,yin2023heterogeneous}. These systems aim to suggest items that align with users' tastes to enchance user engagement. The foundations of the RS can be traced back to collaborative filtering techniques, which leverage user-item interaction data to identify patterns and make recommendations. Over time, these systems haveincorporated more sophisticated approaches, including content-based filtering, hybrid methods, and context-aware recommendations, to address user data's growing complexity and scale.

The development of machine learning and deep learning has revolutionized nearly every field \cite{he2016identity,yang2022retargeting,tian2022changing,tian2024time,koroteev2021bert,koch2020criminal,zhao-etal-2021-structural, ding2024semantic,tao2019fact,tao2024nevlp,zhu2023demonstration}, including recommender systems. These systems now benefit from large-scale models that can leverage vast amounts of data to extract complex relationships. Deep learning techniques, such as neural collaborative filtering (NCF), convolutional neural networks (CNNs), and recurrent neural networks (RNNs), have been employed to enhance the accuracy and robustness of Recommender System \cite{cheng2016wide,he2017neural}. These models benefit from their ability to automatically learn feature representations from raw data, eliminating the need for manual feature engineering and improving predictive performance. Moreover, the use of attention mechanisms and transformer architectures has further advanced the capabilities of deep learning-based recommenders by allowing them to capture sequential and contextual information better \cite{zhou2018deep, pi2020search}.

Recently, Large Language Models (LLMs) like GPT-4 have shown immense potential in understanding and generating human-like text. In recommender systems, vast amounts of natural language data, such as user reviews and product information, are rich in valuable insights \cite{zheng2017joint,li2015online}. 
In this paper, we propose a novel LLM-enhanced deep learning framework with the following contributions:
\begin{itemize}
    \item Developed a framework that efficiently extracts multi-modal information like text and images, from LLMs
    \item Unifiied information from different into same latent space, simplifying the learning process for the ranking model.
    \item Demonstrated how the use of multi-modal information can further enhance the discriminative power of the model, especially for improving false positive rate in the case of the imbalanced dataset. 
\end{itemize}
The structure of the paper is as follows: Section \ref{related} introduces the latest developments in recommender systems and LLMs. Section \ref{method} presents our proposed framework and its key components. Section \ref{experiment} details the experimental setup and analysis. Finally, Section \ref{conclusion} provides concluding remarks.

\section{Related Work}\label{related}

\subsection{Recommender System}
Earlier work on Recommender Systems(RS) did not involve the extensive use of deep learning as seen in current approaches. For specifics, one can refer to~\cite{app7121211}, which includes over 100 techniques from before 2017.

RS can be broadly categorized into personalized~\cite{wu2022knowledge,zheng2022explainable} and group-based~\cite{zan2021uda,gao2024survey} systems. Collaborative Filtering (CF) stands out as a prevalent technique. CF predicts a user’s preferences or opinions by leveraging the collective insights from a large user base. Notable implementations include memory-based CF approaches such as those presented in ~\cite{chen2019collaborative} and ~\cite{barkan2021anchor}, which utilize vector representations. 
In recent years, the integration of graph neural networks like 
GraphSAGE~\cite{hamilton2017inductive}, and others have significantly enhanced model-based CF methods. These models have been extensively applied across various domains, with notable success in music, Point of Interest (POI), and book recommendations \cite{kumar2019predicting} 
\cite{sun2019multi,he2020lightgcn}. 
For instance, the JODIE~\cite{kumar2019predicting} model has been influential in music recommendation, while Multi-GCCF~\cite{sun2019multi}, and LightGCN~\cite{he2020lightgcn} have shown promising results in POI and book recommendation scenarios. Among these, LightGCN has emerged as a classic model in the RS field. The effectiveness of review text in RS has been a subject of debate. For example, ~\cite{chin2018anr} argued that not all parts of reviews hold equal importance, leading them to propose an Aspect-based Neural Recommender (ANR) that focuses on more granular feature representations of items. Similarly, ~\cite{li2019capsule} employed capsule neural networks to extract specific viewpoints and aspects from user and item reviews. Furthermore, ~\cite{wu2019npa} developed a dual-encoder system using CNNs, one for encoding news and the other for learning user profiles based on their interaction with clicked news.

\subsection{Large Language Models Reasoning}
LLMs have demonstrated remarkable reasoning capabilities~\cite{zhao2024selfguidebettertaskspecificinstruction,he2024teacherlmteachingfishgiving}.
Recently, there has been a trend towards using LLMs for traditional tasks. For instance, ~\cite{gptre} employs in-context learning on GPT-3 for Relation Extraction(RE), achieving state-of-the-art (SOTA) performance on multiple test sets. \cite{gptner} adapts LLMs to the Named Entity Recognition (NER) task, aiming to bridge the gap between sequence labeling and text generation. This adaptation demonstrates how LLMs can be fine-tuned or prompted in innovative ways to handle tasks traditionally outside their direct training objectives. ~\cite{zeroshotner} investigate the capabilities of LLMs in zero-shot information extraction scenarios, specifically examining the performance of ChatGPT in the NER task. By focusing on zero-shot learning, the study investigates ChatGPT's ability to identify and classify named entities within text and without any task-specific training data or fine-tuning. ~\cite{gptsignal} conducted the ability of LLMs to generate new financial signals. LLMs have also been employed for other tasks such as text summarization~\cite{goyal2023news}  and sentiment analysis~\cite{sun2023sentiment}.

\subsection{LLM for Recommender Systems}
LLMs have demonstrated remarkable reasoning capabilities~\cite{zhao2024selfguidebettertaskspecificinstruction,he2024teacherlmteachingfishgiving,gptsignal,zhang2024ratt,zhang2024thoughtspaceexplorernavigating}. Recent efforts in the domain of recommender systems have increasingly focused on the utilization of Language Models~\cite{hou2022towards,wang2022towards,yuan2024rhyme}. ~\cite{gao2023chat} utilizes LLMs as the interface for recommender systems, facilitating multi-round recommendations. This enhances both the interactivity and the explainability of the system. ~\cite{wang2023zero} proposed a three-step prompting strategy that substantially surpasses traditional simple prompting techniques in zero-shot settings. ~\cite{wang2023generative} preprocess users’ instructions and traditional feedback, such as clicks, using an instructor module to generate tailored guidance. ~\cite{dai2023uncovering} conduct an evaluation to assess off-the-shelf LLMs for RS, analyzing them from point-wise, pair-wise, and list-wise perspectives

\section{Methodology}\label{method}
Our proposed model leverages deep learning techniques and the advanced reasoning capabilities provided by large language models (LLMs) to enhance the performance of the ranking model. We hypothesize that the summarization power of LLMs can significantly improve the discriminative capabilities of the ranking model, leading to more accurate and relevant recommendations.


\subsection{DLRM and the base model}
DLRM is a robust framework that leverages deep learning techniques for recommendation tasks \cite{naumov2019deep}. DLRM has proven to be highly effective in personalization and recommendation scenarios, such as click-through rate (CTR) prediction.
For a comprehensive understanding of the technical intricacies of Deep Learning Recommendation Models (DLRM), readers are encouraged to consult the original research paper.

In the baseline model, we process the textual information contained in user reviews by converting each text review into an embedding using the sentence transformer. Specifically, we use MiniLM-L6-v2 model for all our experiments \cite{wang2020minilm}. We then take the element-wise average across all dimensions to create user or business features. Similarly, each image contained in the user reviews or associated with a particular business is transformed into continuous data using ResNet50. Specifically, we extract the second-to-last layer to represent the images, capturing rich feature representations.

\subsection{LLM summarization}

To leverage the summarizing power of large language models (LLMs), we propose various methods to enhance the features fed into our model. In this section, we explain how the LLM-enhanced DLRM differs from the base model presented above.
\begin{itemize}
    \item Dense Features: In addition to the continuous features used in the base model (e.g., the number of reviews a restaurant received, average rating), we utilize the LLM's reasoning ability to extract pricing information from user reviews. This enriched feature set provides a more comprehensive understanding of the restaurant being scored.
    \item Sparse Features: Instead of applying an element-wise average across embeddings from all textual reviews or images, we use the LLM to process and summarize the most important information from all reviews, obtaining an embedding for this summary alone. For images, we leverage the LLM's multimodal capabilities to interpret and summarize the information contained in the images, converting them into textual descriptions. This textual information is then processed in a similar manner to the reviews.
\end{itemize}

This approach offers several advantages:
\begin{itemize}
    \item Reduction of Noise: By summarizing the most important information, we ensure that only relevant data is fed into the model, preventing irrelevant or noisy information from diluting important signals.
    \item Unified Embedding Technique: Since images are converted into textual descriptions, both reviews and images use the same embedding technique to transform them into continuous data. This ensures that features from different modalities are projected into the same latent space, enhancing the model's ability to understand and utilize the combined information effectively.
\end{itemize}

\begin{figure*}[h!]
    \centering
    \includegraphics[width = \textwidth,  height= 0.3\textheight]{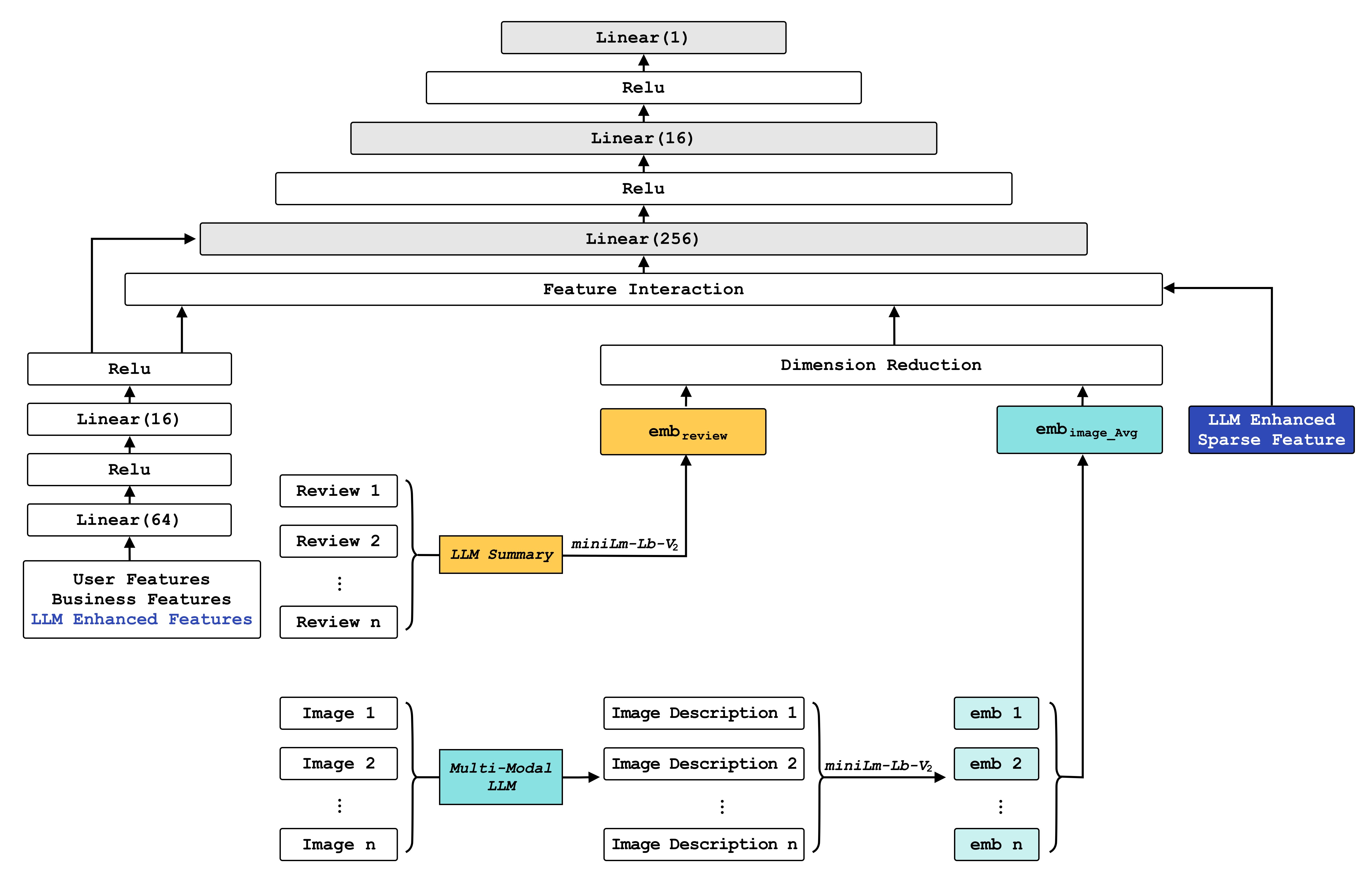}
    \caption{LLM enhanced DLRM for the restaurant recommendation task}
    \label{proposed_model}
\end{figure*}

\begin{figure}[hbt!]
    \centering
    \includegraphics[width=\linewidth, height=0.3\textheight]{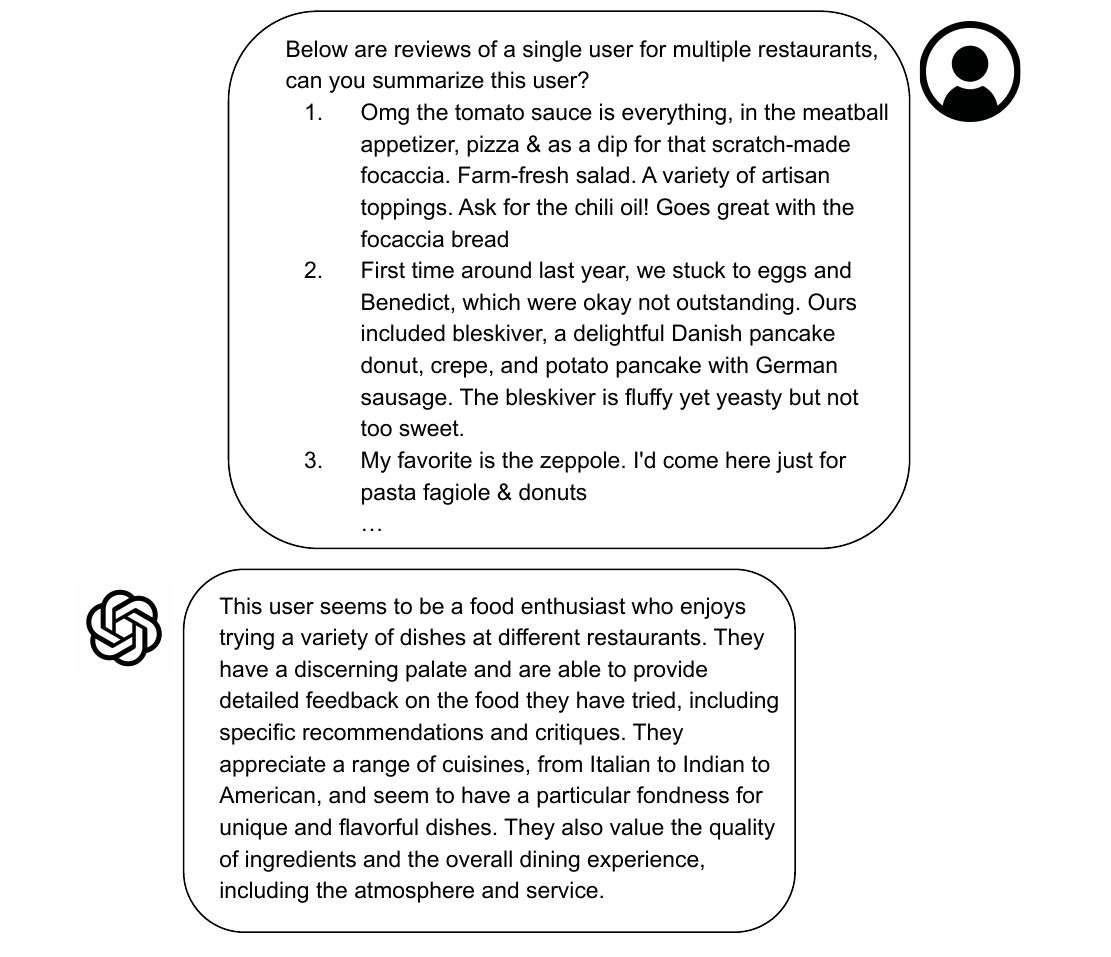}
    \caption{LLM user summary example.}
    \label{fig:llmsummary}
\end{figure}

Besides the differences mentioned above, as shown in Figure \ref{proposed_model}, we also introduce an additional sparse feature into the model. Using the LLM, we categorize each restaurant into one of 11 categories. This categorical feature is then fed into the feature interaction module after the embedding layer.
\subsection{Dimension reduction}
To mitigate the risk of overfitting caused by the high-dimensional outputs of both the sentence transformer (384 dimensions) and ResNet50 (2048 dimensions), we propose an upstream model for dimensionality reduction. This approach preserves meaningful information while addressing the potential increase in model parameters.
Our method involves the following steps:
\begin{itemize}
    \item Concatenate embeddings from both the text encoder and image encoder into a single tensor.
    \item Feed the tensor through a Multi-Layer Perceptron (MLP).
    \item The MLP outputs the probability of the outcome of interest (e.g., a positive review).
\end{itemize}
Importantly, we apply the same MLP used in training to the embeddings during the testing phase. This approach ensures effective dimensionality reduction while retaining the most crucial information for prediction. By implementing this technique, we balance model complexity and predictive power, enhancing the overall the recommender system performance.

\section{Experiement}\label{experiment}
In this study, we utilized a comprehensive dataset tailored for restaurant reviews analysis. This data is published in Kaggle\footnote{\url{https://www.kaggle.com/}} and it is collected from Google reviews~\cite{googlerestaurantreviews}. This dataset comprises user-generated reviews for various restaurants. 
\subsection{Parameter and Configuration}
To evaluate the model's performance under different conditions, We experimented with various dropout rates [0.1, 0.3, 0.5] and applied different weighted loss functions (basic and square root) to address data imbalance.
    \begin{itemize}
        \item Basic: \begin{equation}
                        W^{label} = 1 - \frac{Num^{label}}{Num^{total}}
                        \end{equation}
                        
        \item Square root: 
                        \begin{equation}
                        W^{label} = \sqrt{1 - \frac{Num^{label}}{Num^{total}}}
                        \end{equation}
    \end{itemize}

We used a learning rate of 0.01 and applied early stopping after 300 epochs when the false positive rate showed no improvement for 50 epochs. We repeated each parameter set evaluation five times.

\begin{table}[h]
\caption{Performance of best model against training set}
\small  %
\begin{tabularx}{\linewidth}{Xlllll}
        \toprule
         &  &  & accuracy & fp rate & loss \\
        model & wgted loss & dropout &  &  &  \\
        \midrule
        \multirow[t]{6}{*}{proposed} & \multirow[t]{3}{*}{basic} & \textbf{0.10} & \textbf{91.62\%} & \textbf{2.02\%} & 2.7 \\
         &  & 0.30 & 88.11\% & 4.14\% & 4.08 \\
         &  & 0.50 & 86.06\% & 7.06\% & 5.28 \\
        \cline{2-6}
         & \multirow[t]{3}{*}{sqrt root} & 0.10 & 95.95\% & 4.36\% & 4.36 \\
         &  & 0.30 & 94.14\% & 13.69\% & 6.35 \\
         &  & 0.50 & 91.95\% & 18.87\% & 9.19 \\
        \cline{1-6}
        \multirow[t]{6}{*}{baseline} & \multirow[t]{3}{*}{basic} & 0.10 & 92.51\% & 26.58\% & 6.35 \\
         &  & 0.30 & 92.39\% & 27.82\% & 6.82 \\
         &  & 0.50 & 92.23\% & 28.27\% & 6.97 \\
        \cline{2-6}
         & \multirow[t]{3}{*}{sqrt root} & 0.10 & 94.12\% & 29.01\% & 10.46 \\
         &  & 0.30 & 93.97\% & 31.02\% & 11.07 \\
         &  & 0.50 & 94.06\% & 33.49\% & 11.54 \\
         \cline{1-6}
         \multirow[t]{6}{*}{proposed-text\footnotemark[1]} & \multirow[t]{3}{*}{basic} & 0.10 & 94.69\% & 0.7\% & 1.74 \\
 &  & 0.30 & 87.78\% & 2.68\% & 3.76 \\
 &  & 0.50 & 83.2\% & 4.31\% & 5.35 \\
\cline{2-6}
 & \multirow[t]{3}{*}{sqrt root} & 0.10 & 97.38\% & 3.67\% & 2.88 \\
 &  & 0.30 & 93.98\% & 8.95\% & 6.09 \\
 &  & 0.50 & 92.99\% & 24.56\% & 8.65 \\
         \cline{1-6}
    \multirow[t]{3}{*}{proposed-image\footnotemark[2]} & \multirow[t]{3}{*}{basic} & 0.10 & 69.32\% & 8.77\% & 8.14 \\
         &  & 0.30 & 63.52\% & 12.38\% & 9.45 \\
         &  & 0.50 & 60.33\% & 18.25\% & 10.51 \\
        \cline{2-6}
         & \multirow[t]{3}{*}{sqrt root} & 0.10 & 90.26\% & 54.93\% & 13.68 \\
         &  & 0.30 & 89.47\% & 62.06\% & 16.21 \\
         &  & 0.50 & 88.57\% & 71.56\% & 18.22 \\         
        \bottomrule
    \end{tabularx}
\end{table}

\begin{table}[h]
\caption{Performance of best model against testing set}
\small  %
\begin{tabularx}{\linewidth}{Xlllll}
\toprule
 &  &  & accuracy & fp rate & loss \\
model & wgted loss & dropout &  &  &  \\
\midrule
\multirow[t]{6}{*}{proposed} & \multirow[t]{3}{*}{basic} & 0.10 & 85.27\% & 27.22\% & 50.18 \\
 &  & 0.30 & 84.02\% & 20.84\% & 30.66 \\
 &  & \textbf{0.50} & \textbf{83.48\%} & \textbf{18.16\%} & 15.96 \\
\cline{2-6}
 & \multirow[t]{3}{*}{sqrt root} & 0.10 & 88.05\% & 38.68\% & 66.51 \\
 &  & 0.30 & 89.15\% & 36.42\% & 49.68 \\
 &  & 0.50 & 89.02\% & 31.82\% & 23.76 \\
\cline{1-6}
\multirow[t]{6}{*}{baseline} & \multirow[t]{3}{*}{basic} & 0.10 & 91.36\% & 31.80\% & 9.32 \\
 &  & 0.30 & 91.89\% & 31.74\% & 7.66 \\
 &  & 0.50 & 91.83\% & 31.79\% & 7.64 \\
\cline{2-6}
 & \multirow[t]{3}{*}{sqrt root} & 0.10 & 92.96\% & 35.26\% & 14.57 \\
 &  & 0.30 & 93.26\% & 35.66\% & 12.62 \\
 &  & \textbf{0.50} & \textbf{93.47\%} & \textbf{37.58\%} & 12.59 \\
 \cline{1-6}
\multirow[t]{6}{*}{proposed-text\footnotemark[1]} & \multirow[t]{3}{*}{basic} & 0.10 & 84.17\% & 38.68\% & 63.91 \\
&  & 0.30 & 81.59\% & 26.8\% & 36.03 \\
&  & 0.50 & 79.1\% & 19.09\% & 24.16 \\
\cline{2-6}
& \multirow[t]{3}{*}{sqrt root} & 0.10 & 86.53\% & 48.84\% & 88.24 \\
&  & 0.30 & 87.01\% & 40.9\% & 50.87 \\
&  & 0.50 & 88.15\% & 44.66\% & 36.05 \\
 \cline{1-6}
 \multirow[t]{6}{*}{proposed-image\footnotemark[2]} & \multirow[t]{3}{*}{basic} & 0.10 & 59.8\% & 38.53\% & 34.67 \\
 &  & 0.30 & 58.15\% & 32.68\% & 25.8 \\
 &  & 0.50 & 57.84\% & 28.3\% & 21.97 \\
\cline{2-6}
 & \multirow[t]{3}{*}{sqrt root} & 0.10 & 84.92\% & 76.23\% & 49.11 \\
 &  & 0.30 & 85.79\% & 76.08\% & 37.52 \\
 &  & 0.50 & 86.53\% & 79.72\% & 28.67 \\
\bottomrule
\end{tabularx}
\end{table}

\footnotetext[1]{This model is based on the proposed model but for the embedding features, retains only text review}
\footnotetext[2]{ This model is based on the proposed model but for the embedding features, retains only image review}
\subsection{Data Pre-processing}

Our baseline model utilizes dense features and embedding features to predict the review rating, where we exclude the current review to avoid bias in generating user and business features, ensuring that the current user's or business's review does not influence the feature construction. Weuse the entire training dataset to construct testing features.
\begin{itemize}
    \item Dense Features: 
    \begin{itemize}
        \item Total number of reviews received by this business.
        \item Average rating of all reviews by this user 
        \item Average rating of this business
    \end{itemize}
    \item Embedding Features:
    \begin{itemize}
        \item User Review Text Feature: All review written by the same user are independently converted into 384-dimensional embedding vectors through sentence transformer, then apply average pooling to get a single 1x384 vector.
        \item Business Review Text Feature: Similarly, process reviews for same business into 1 x 384 vector.
        \item Review Image Feature: Firstly, input image into a pre-trained reset-50 model, then extract features from the second-to-last layer. Each image generates a 1x2048-dimensional embedding vector then apply average pooling to single vector.
        \item Finally, we concatenate these 3 features, process with a upstream model to reduce its dimension to 32.
    \end{itemize}
\end{itemize}

Compared to the baseline model, we add new dense feature: Price Tag, and a sparse Feature: Restaurant Category, into the proposed model, transform all embedding features through prompting engineering with GPT 3.5-turbo-1106 model and multi-modal model.

\begin{itemize}
    \item Dense Features:
    \begin{itemize}
        \item \textbf{Price Tag Feature}: Use the prompt: “Can you tell me if the price is over-rice, fair price, low price from reviews for this restaurant. Give me just the category ”.  With post-processing, we generate a price tag categorized as fair price, overpriced, cheap price, or none (if no clean indication of price level in reviews). 
    \end{itemize}

    \item Sparse Features:
    \begin{itemize}
        \item \textbf{Restaurant Category Feature}: Use prompt \textit{“can you tell me what kind of restaurant this is from these reviews for the restaurant. Return me in this format:’{type}’”}. 
        There are 179 distinct types and the maximum number of subtypes for a restaurant is 11. In the pre-processing step, all subtype tensors were padded to the length of 11, and the padding value is set to 179 (i.e. embedding table contains 180 distinct values and the last one is padding idx).
    \end{itemize}

    \item Embedding Features with LLM:
    \begin{itemize}
        \item Get a summary of review with prompts shown in  Fig~\ref{fig:llmsummary} to summarize all reviews written by the same user, then convert the summary into an embedding vector representing the user. Similarly, get the embedding vector for a single business. We concatenate the these two review summary features into one big vector, then passed through an upstream model to reduce vector dimensionality to 32.
        \item We rely on a multi-modal model (BLIP 2) to produce one description sentence for images through unconditional image captioning as shown in Fig~\ref{fig:blip2example}. Transformed sentence into an embedding vector, then apply average pooling on all images' vecotr and get one vector, which is finally passed through an upstream model to reduce its dimensionality to 32.
        \item Concatenating the text and image vector 
    \end{itemize}
\begin{figure}[hbt!]
    \centering
    \includegraphics[width=0.8\linewidth]{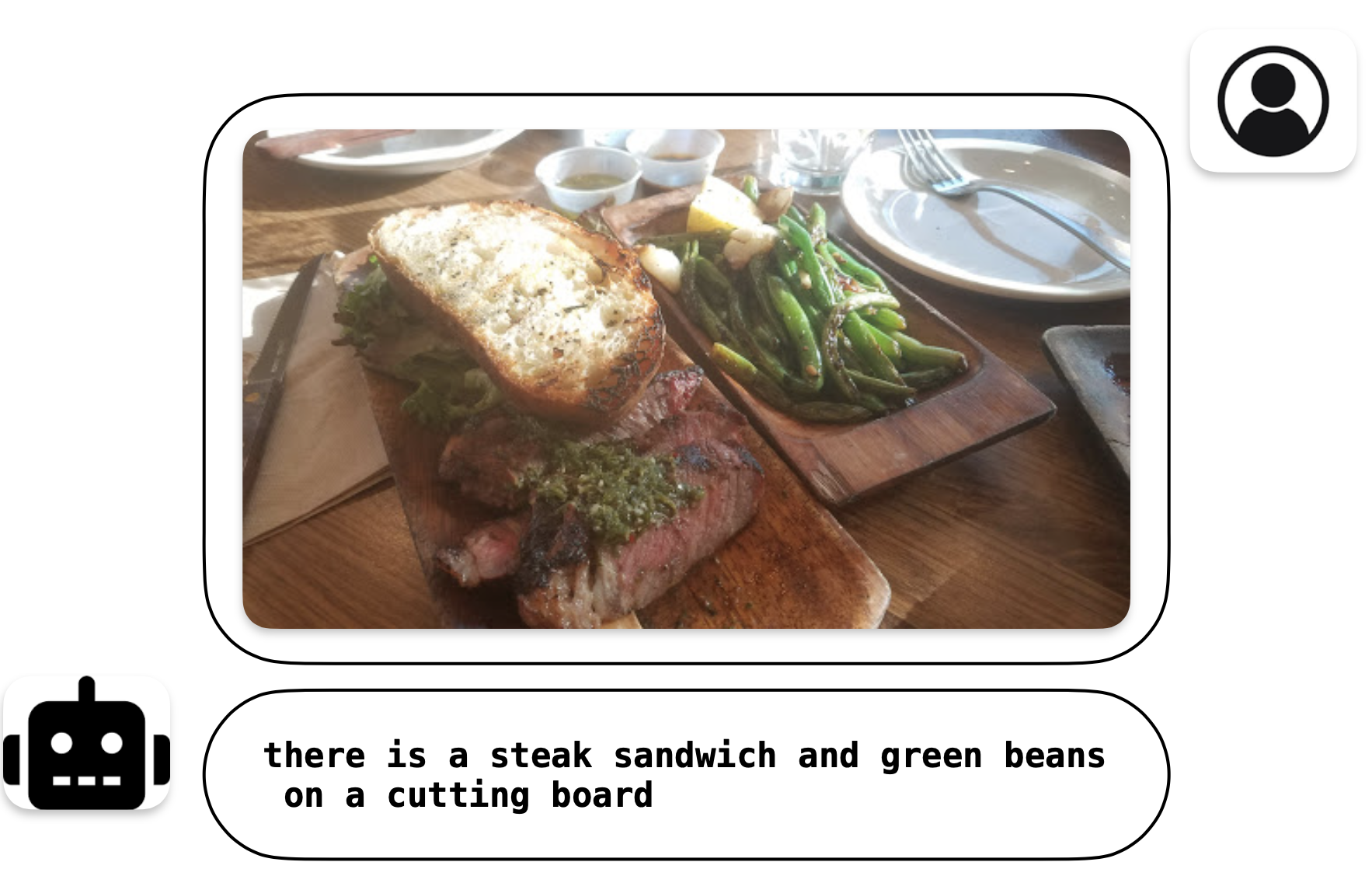}
    \caption{Using BLIP-2 to perform unconditional image captioning on restaurant review image}
    \label{fig:blip2example}
\end{figure}
\end{itemize}

\subsection{Result}
\begin{itemize}
    \item Compared to the baseline model, the proposed model achieves a much better false positive rate in both the train set and the test set as shown in Table 1. The best training set false positive rate is around 2\%.
    \item Given Table 2, the best model for test set false positive rate is the proposed model with basic weighted loss and 0.5 dropout rate, achieving 18.16\% false positive rate while accuracy is 83.48\% {\bf  we obtain a 19.4\% improvement in the false positive rate at the expense of 10\% decrease in accuracy}
    \item A comprehensive ablation study is conducted to investigate the efficacy of different modalities in proposed model. The results revealed a notable regression in both overall accuracy and false positive rate, with the latter showing a more pronounced decline. This finding suggests that both image and text modalities contribute significantly to enhancing the model's overall performance by providing distinct signals. These modalities prove particularly valuable in determining the alignment between items and user interests. The synergistic effect of combining visual and textual information appears to be crucial in reducing false positives and improving the model's discriminative power.
\end{itemize}
\subsection{Analysis}


\sloppy
In the context of ranking and recommendation, a high false positive rate is unacceptable. In most practical RS, top 1 accuracy becomes less important and top N accuracy is high enough. A lower false positive rate ensures that the recommendations are more aligned with the users' tastes. The baseline model focuses on accuracy due to imbalanced data but struggles with false sample identification, making the proposed model's lower false positive rate more valuable.

The significant reduction in the false positive rate observed in the proposed model can be attributed to the powerful ability of Large Language Models (LLMs) in summarizing reviews. 
In addition, LLMs are good at extracting and emphasizing critical and repeated information in various reviews. This summarization power ensures that the model captures the essential features that distinguish different users and restaurants, while, the baseline model take average of the embedding vectors. This method tends to dilute the information because it treats all reviews equally, regardless of their quality or relevance, hence introduce more nosie and halt performance.

The proposed model leverages the multimodal model (BLIP2) and its description ability to identify multiple food items, help to discriminate between various users and restaurants. By contrast, the baseline model relies on the image classification model(resnet), which has limitations in identifying multiple objects within the image, especially when dealing with multiple types, and fail to capture the semantic meaning within images, making the extracted signals less powerful.

 Our ablation study shows that both LLM and multimodal features provide critical information for predicting user preferences. Removing either modality results in a performance drop, indicating that both text and image reviews contribute valuable insights
\section{Conclusion}\label{conclusion}
In this paper, we have proposed an innovative framework that harnesses the reasoning and summarization capabilities of LLMs to process multi-modal information effectively, demonstrating the significant potential of integrating multi-modal data to enhance the performance of deep learning-based recommender systems, particularly in scenarios involving imbalanced datasets.

This novel method convert image-based information into textual data, allowing both to be processed by same text encoder \. Consequently, both image-derived and text-based features are represented in the same latent space, ensuring a more cohesive and comprehensive input.

Our findings show that LLM-generated signals greatly improve model performance, where enhancement is mainly driven by two factors:
\begin{itemize}
    \item The ability to extract valuable insights from negative reviews, which often contain critical information for recommendation systems.
    \item LLMs distill essential information, avoiding the dilution common in traditional averaging approaches.
\end{itemize}



The improved performance underscores the potential of LLMs and multi-modal models in enchancing recommender systems. More accurate and effective recommendations are achieved by leveraging the contextual understanding and discriminative capabilities of these models.

\bibliographystyle{IEEEtran}
\bibliography{IEEEexample}
\vspace{12pt}

\end{document}